\documentclass{article}
\usepackage{spconf,amsmath,graphicx}

\usepackage{multicol, multirow}
\usepackage{adjustbox}
\usepackage{amssymb}
\usepackage{mathtools}
\usepackage{makecell, booktabs}
\usepackage{url, cite}
\usepackage{tabularx}
\newcolumntype{Y}{>{\centering\arraybackslash}X}
\usepackage{hyperref}
\usepackage{arydshln}
\usepackage{booktabs}
\hypersetup{
    colorlinks=true,
    linkcolor=purple,
    filecolor=magenta,      
    urlcolor=purple,
}
\usepackage{xcolor}

\title{attentive max feature map and joint training \\ for acoustic scene classification}
\name{Hye-jin Shim$^{1}$, Jee-weon Jung$^{2}$, Ju-ho Kim$^{1}$, and Ha-Jin Yu$^{1*}$\thanks{$^*$Corresponding author.}}
\address{$^1$School of Computer Science, University of Seoul, $^2$Naver Corporation}
\begin{document}
\ninept
\maketitle

\begin{abstract}
Various attention mechanisms are being widely applied to acoustic scene classification. 
However, we empirically found that the attention mechanism can excessively discard potentially valuable information, despite improving performance. 
We propose the attentive max feature map that combines two effective techniques, attention and a max feature map, to further elaborate the attention mechanism and mitigate the above-mentioned phenomenon. 
We also explore various joint training methods, including multi-task learning, that allocate additional abstract labels for each audio recording.
Our proposed system demonstrates state-of-the-art performance for single systems on Subtask A of the DCASE 2020 challenge by applying the two proposed techniques using relatively fewer parameters.
Furthermore, adopting the proposed attentive max feature map, our team placed fourth in the recent DCASE 2021 challenge. 
\end{abstract}

\section{Introduction}
\label{sec:intro}
Acoustic scene classification (ASC) is the task of recognizing scenes based on environmental sounds.
The detection and classification of acoustic scenes and events (DCASE) community hosted several challenges \cite{mesaros2018multi, mesaros2019acoustic, heittola2020acoustic}. 
The DCASE 2020 challenge included two subtasks addressing different properties for ASC: 1) Subtask A requires the generalization to unknown devices, and 2) Subtask B demands a low-complexity solution in terms of model size (i.e., the number of parameters).
Subtask A aims to identify a given audio clip recorded using multiple devices into one of the ten predefined acoustic scenes including the pedestrian street, airport, and tram.
Subtask B aims to classify a given audio clip into one of the three relatively abstract level classes: outdoor, indoor, and transportation.
The classification task for subtask A and subtask B are referred to as 10-class and 3-class classifications.

\begin{figure*}[ht!]
\begin{center}
    \centering
    \includegraphics[width=0.85\linewidth]{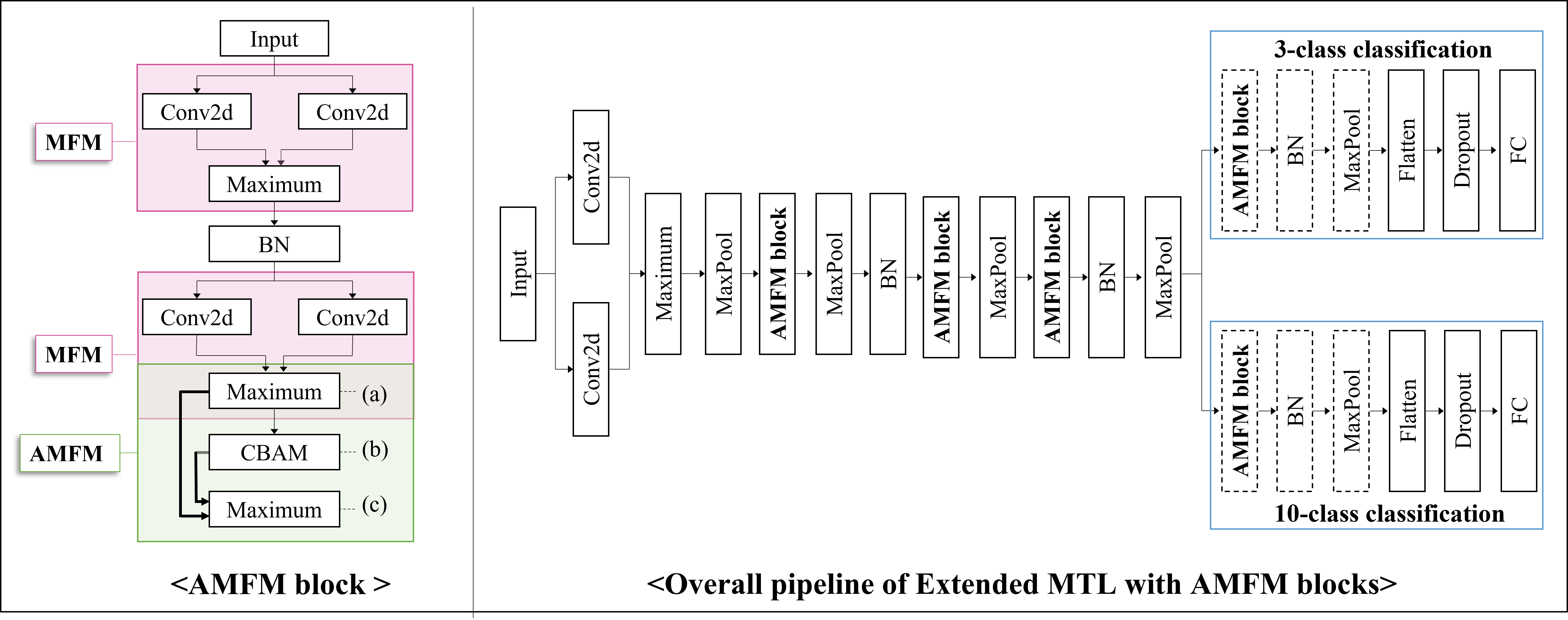}
\vspace{-1em}
\caption{Proposed attentive max feature map (AMFM) block structure (left) and overall pipeline of the extended multi-task learning (MTL) structure  (right). 
The AMFM block on the left comprises both the max feature map (MFM) and AMFM, illustrated in pink and green boxes, respectively. 
The illustrated overall pipeline adopts the AMFM block and demonstrates the best results among the explored joint training methods.
}
\label{fig:architecture}
\end{center}
\vspace{-1.5em}
\end{figure*}

Recent studies in ASC can be primarily divided into two strands, data preprocessing and modeling, and most top-ranking systems focused on data preprocessing in the DCASE 2020 challenge. 
In data preprocessing, most recent systems have exploited delta and delta-deltas \cite{suh2020designing, hu2020device, gao2020acoustic, liu2020acoustic, koutini2020cp}, diverse data augmentation techniques (e.g., mixup \cite{zhang2018mixup}, and SpecAugment \cite{park2019specaugment}), sub-frequency analysis \cite{phaye2019subspectralnet, suh2020designing}, feature investigation, and temporal division \cite{mun2017deep}. 
In modeling, ResNet \cite{he2016deep} and convolutional neural networks (CNNs) are the most widely used architectures for ASC tasks. 
Techniques such as the adoption of the knowledge distillation framework \cite{heo2019acoustic, jung2020knowledge} and receptive field regularization \cite{koutini2020cp} have been demonstrated as effective. 
Utilizing the representation of other tasks and training a general-purpose network have also been investigated \cite{shim2020audio, jung2020acoustic, kong2018dcase, jung2021dcasenet}.
Among various modeling approaches, the attention mechanism is one of the most effective techniques \cite{ren2019attention, phan2019spatio}.

In this paper, we propose two modeling techniques independent of data preprocessing: an attentive max feature map (AMFM) and joint training of the concrete classes of Subtask A and abstract classes of Subtask B. 
First, by visualizing the effect of attention on the feature map (see Figure~\ref{fig:featuremap}), applying attention excessively discards potentially important information, despite improving the performance.
Analyzing the visualization, we assume that preserving more information might further boost the discriminative power of a feature if adequately addressed. 
This assumption is in line with recent studies that have adopted sigmoid-based attention to alleviate discarding too much information \cite{woo2018cbam}. 
We argue that the max feature map (MFM), a technique that adopts a competitive scheme via elementwise max operation~\cite{wu2018light}, can be used for this purpose because MFM showed its effectiveness in~\cite{shim2020audio, shim2020capturing, lee2021cnn}. 
Specifically, we design a new technique combining attention and MFM, which we refer to as the AMFM, in which attention emphasizes the most informative region and the MFM prevents excessive information loss using a max operation.
The proposed AMFM compares the feature maps before and after the attention mechanism to mitigate excessive information loss.

Second, we propose a joint training scheme using the 10-class label from Subtask A and 3-class label from Subtask B to improve the performance of Subtask A. 
We explore four methods for training the two tasks.
The underlying assumption is that additional labels can improve the supervision in the training process. 
The proposed approach is similar to the work by Hu et al. \cite{hu2020device, hu2021two}, which employed the prediction of the classifiers of Subtask B to improve performance on Subtask A. 
The two classifiers were trained separately by Hu et al., whereas the proposed approach extends the framework by training a single model with the joint training scheme.
To the best of our knowledge, this approach is the first to simultaneously train a single ASC system using labels for both Subtasks A and B.
The performance of the proposed system is comparable to that of the state-of-the-art approaches without complex data preprocessing techniques.
Besides, our team could take fourth place in the DCASE 2021 challenge adopting the proposed AMFM.

\section{Proposed methods}
\subsection{Attentive max feature map}
\label{sec:format}

The MFM replaces the non-linear activation function, typically the rectified linear unit (ReLU), with a competitive scheme~\cite{wu2018light}.
It was originally proposed for situations where noisy labels are dominant. 
Wu et al. argued that using threshold-based non-linearity, such as ReLU, might not effectively generalize on unknown data distributions because an activation function separates noisy and informative signals using a threshold (or bias) learned from training data. 
Using threshold-based non-linearity might cause information loss, especially for the first few convolutional layers~\cite{wu2018light}.
To overcome this issue, MFM adopts an elementwise max operation in place of ReLU non-linearity.
The max operation selects relatively important features learned by different filters.
Existing studies~\cite{shim2020audio, shim2020capturing, lee2021cnn} have demonstrated that MFM is effective in ASC.

The implementation of the MFM operation can be described as follows. 
Let $a$ be an output feature map of a convolutional layer, $a \in \mathbb{R}^{K \times T \times F}$, where $K$, $T$, and $F$ refer to the number of output channels, time-domain frames, and frequency bins, respectively. 
In addition, $a$ is first split into two equal-sized feature maps, $a_1$ and $a_2$, where $a_1$, $a_2$ $\in \mathbb{R}^{\frac{K}{2} \times T \times F}$. 
The MFM applied feature map is obtained by applying $Max(a_1, a_2)$ elementwise. 
On the left in Figure~\ref{fig:architecture}, the pink box describes the conventional MFM operation.

\begin{figure*}[ht!]
\begin{center}
    \centering
    \includegraphics[width=0.85\linewidth]{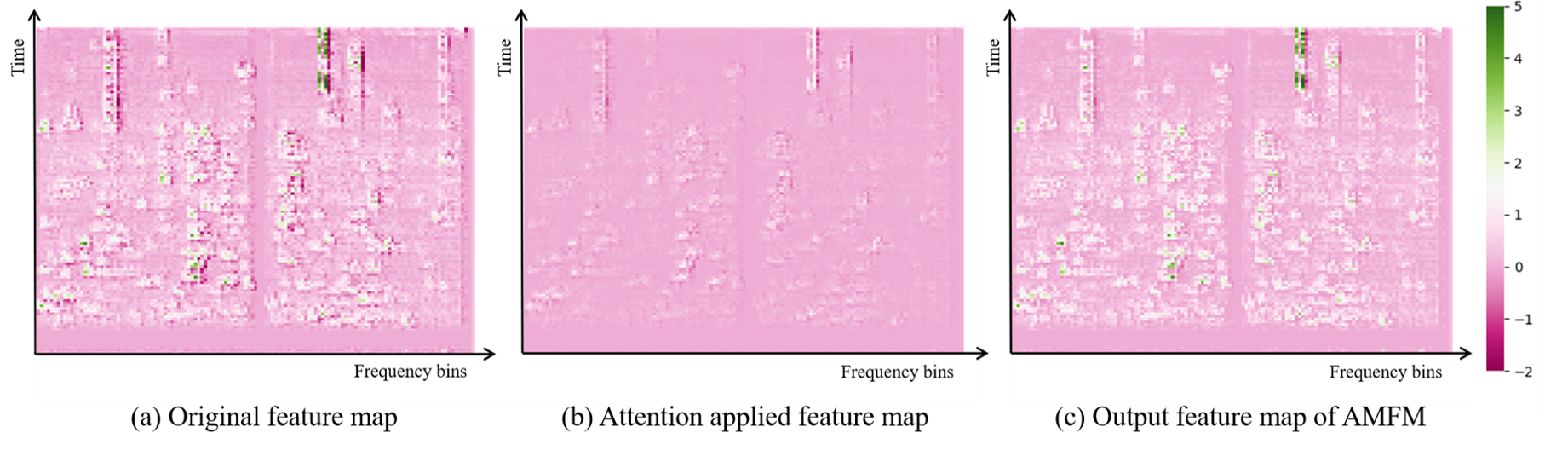}
\vspace{-1.5em}
\caption{Each feature map indicates (a) before and (b) after the attention mechanism, and (c) the output of the attentive max feature map (AMFM). These were extracted from the first AMFM block at positions (a)-(c) on the left in Figure~\ref{fig:architecture}. 
This demonstrates that the attention mechanism can cause excessive information loss. 
After applying the AMFM (c) with a comparison of the two feature maps of (a) and (b), the features were found to enrich the representation preventing information loss (Best viewed in color.).}
\label{fig:featuremap}
\end{center}
\vspace{-1.5em}
\end{figure*}

The recent literature on the ASC task has demonstrated the effectiveness of the attention mechanism~\cite{ren2019attention, phan2019spatio, liu2020acoustic}.  
The attention mechanism highlights important information and enriches the representation. 
Among various attention mechanisms, the convolutional block attention module (CBAM) \cite{woo2018cbam} considers both channel and spatial attention, and has the advantage of seamless implementation regardless of the architecture.
Our previous work confirmed that MFM and CBAM can be employed simultaneously~\cite{shim2020capturing}.
However, through an analysis using the visualization of intermediate feature maps, we empirically found that information is excessively discarded, emphasizing only a small fraction (see Figure~\ref{fig:featuremap} (b)).

We hypothesize that, although the attention mechanism is an effective technique, preserving relatively more information might further improve the discriminative power of a feature. 
Thus, we argue that combining the two existing techniques (the attention mechanism and MFM) might leverage the effectiveness of the attention mechanism while restricting excessive information deletion. 
The proposed technique based on this inspiration is the AMFM, which competitively applies the attention mechanism. 
It compares two feature maps before and after the attention mechanism and outputs their elementwise maximum values as Figure~\ref{fig:featuremap} shows.

The proposed AMFM performs $Max(a_1, CBAM(a_1))$ elementwise. 
We illustrate an AMFM block that involves both MFM (pink box) and AMFM (green box) on the left in Figure~\ref{fig:architecture}.
Conv2d, Maximum, and BN refer to the 2D convolutional layer, maximum operation, and batch normalization, respectively.
In the AMFM block, (a), (b), and (c) in Figure~\ref{fig:architecture} are consistent with those in Figure~\ref{fig:featuremap}. 
Figure~\ref{fig:featuremap} indicates that AMFM alleviates exaggerated attention and selects salient representation as intended.
Moreover, AMFM can be applied to architectures where various attention mechanisms are used, although we combined MFM with the CBAM attention mechanism.

\subsection{Joint training}
When the tasks are highly related, simultaneously training them using an integrated model provides better supervision as reported by~\cite{zamir2018taskonomy,dwivedi2019representation}. 
Thus, in this particular case of the ASC task, we assumed that adopting multi-task learning (MTL) using the two labels (3-class and 10-class) would be helpful. 
These two labels of the two subtasks only differ in the degree of abstraction, where a hierarchy exists (each class in the 3-class definition is further divided into three or four classes in the 10-class definition). 
Therefore, we propose to jointly train the model using both labels. 

The term ``joint training'' in this paper includes pre-training, original MTL, and variants of MTL. 
The study by Hu et al.~\cite{hu2020device} is one of the most similar studies on ASC that considers the relationship between the two subtasks. 
The difference between our work and \cite{hu2020device} is that \cite{hu2020device} employed joint prediction where the final prediction is performed using the score fusion of the two separately trained classifiers. 
In contrast, our work trains an integrated model.

\begin{table}[t]
  \caption{Performance comparison according to the application of the attention mechanism and the structure of the convolutional neural network (CNN), max feature map (MFM), and attentive max feature map (AMFM).} 
  \label{tab:amfm}
  \centering
  \begin{tabular}{lcc}
    \toprule
    System & Attention & Acc (\%) \\
    \midrule
    \multirow{2}{*}{CNN w/ ReLU} & X &70.2 \\
     & O &68.3 \\
    \midrule
    \multirow{2}{*}{CNN w/ LeakyReLU} & X &69.6 \\
     & O &68.2 \\ 
    \midrule
    \multirow{2}{*}{MFM} & X & 69.4 \\
     & O & 70.4 \\
    \midrule
    \textbf{AMFM} & O & \textbf{70.8} \\
    \bottomrule
  \end{tabular}
  \vspace{-5mm}
\end{table}

We explore four different methods for joint training of two subtasks where the first two methods directly apply the existing methods without modification, and the other two methods alter the MTL framework.
First, we adopt the pre-training method. 
The system is first trained for the 3-class classification and then fine-tuned with the 10-class classification.
Second, we apply the original MTL \cite{caruana1997multitask} for the two subtasks.
In this case, the network is designed to learn the shared representation, and the last hidden layer is directly connected with the output layers of each task.

Third, we exploit the extended MTL architecture, which has additional layers allocated for each training task, after the last hidden layer shared by the two tasks. 
Finally, we investigate the sequential order of training for the two tasks, considering the hierarchical relationship between the two tasks. 
Unlike the original MTL, the two classifiers are not connected to the same layer. 
Instead, the proposed sequential MTL has a hierarchical design, where the output layer for the 3-class classification is placed after an intermediate hidden layer. 
This architecture design is in line with that proposed by~\cite{jimenez2018sound}. 
Comparing the order of the 3-class and 10-class classifications, training the 3-class classification first, followed by the 10-class classification is more effective. 
This finding is similar to the majority of deep learning structures that deal with abstract representations in the hidden layers close to the input and specific/sophisticated representations in the hidden layer close to the output layer~\cite{maninis2019attentive}.

For further improvement, we additionally explore the joint prediction proposed in \cite{hu2020device} for the extended and sequential MTL framework. 
Joint prediction exploits the score fusion of 3-class and 10-class classifications, where the 3-class classification result is used as prior knowledge.
In terms of adjusting the weight ratio between the two tasks, we explored both intuitive and methodological approaches: a grid search and GradNorm~\cite{chen2018gradnorm}, respectively. 

\section{Experimental settings and results}
\subsection{Experimental settings}
All experiments in this paper used the DCASE 2020 Task 1 Subtask A (1-A) dataset with the corresponding 3-class labels from Subtask B.
We used 13,965 audio clips and 2,970 audio clips for training and evaluating the model, respectively, following the official protocol.
The DCASE 2020 Task 1-A dataset consists of various audio clips collected from three real devices (A, B, and C) and six simulated devices (s1 to s6).
Only devices A-C and s1 to s3 were used in the training set, and s4 to s6 were unavailable in the training phase. 
Each audio clip has a duration of 10s with a 44.1kHz sampling rate and 24-bit resolution. 
We used 256-dimensional Mel-spectrograms as the base feature.
A short-time Fourier transform with 2,048 FFT points was applied with a 40 ms window size and 20 ms hop length. 
Mixup \cite{zhang2018mixup} and SpecAugment \cite{park2019specaugment} were exploited for data augmentation.
The initial learning rate was set to 0.001 and scheduled with a warm restart of the stochastic gradient descent (SGD).
The SGD optimizer with a momentum of 0.9 was used. 
The batch size and number of epochs were set to 24 and 800, respectively.
The architecture details are similar to those described in \cite{shim2020capturing}.
In the case of additional blocks for the MTL, the parameters of the AMFM block are identical to those for the other blocks.
The last hidden layer for each task has 100 nodes followed by the output layer for each subtask. 
All models were implemented in PyTorch, a deep learning library in Python, and the code will be publicly available~\footnote{URL to be made available upon publication}. 
Other data preprocessing techniques, such as the application of logarithms, deltas, double-deltas, and sub-band frequency separation, were not used in this work, which leaves room for further improvements. 

\subsection{Result analysis}

\subsubsection{Attentive max feature map}
Table \ref{tab:amfm} delivers the results of comparing the effectiveness of applying the CBAM in various deep neural network structures: CNN, MFM, and AMFM.
In the experiments, CBAM decreased performance when applied to a CNN with  the ReLU or leaky ReLU activations but improved performance when applied using MFM.
Using the proposed AMFM, the best performance of 70.8\% accuracy was achieved. 
By comparatively analyzing (a), (b), and (c) in Figure~\ref{fig:featuremap} jointly with their corresponding accuracies (70.2\%, 68.3\%, and 70.8\%, respectively), we argue that AMFM emphasizes discriminative information while avoiding excessive information removal.

\begin{table}[t]
  \caption{Comparison of various joint training strategies.}
  \label{tab:joint}
  \centering
  \begin{tabular}{lccc}
    \toprule
    \multirow{2}{*}{System}  & Joint & \multirow{2}{*}{\# Params}  & \multirow{2}{*}{Acc (\%)}  \\
    & prediction  &  & \\
    \midrule
    w/o joing training & X & 1.5M & 70.8\\
    \midrule
    Pre-traing & X & 1.5M & 69.2\\
    Conventional MTL & X & 1.5M & 69.7\\
    \midrule
    \multirow{2}{*}{Extended MTL} & X & 0.6M & \textbf{71.3} \\
     & O & 0.6M & 70.0 \\
    \midrule
    \multirow{2}{*}{Sequential MTL} & X & 0.7M & 71.0 \\
     & O & 0.7M & 69.1\\
    \midrule
    Separated Classifier \cite{hu2020device} & O & 1.5M & 69.4 \\
    \bottomrule
  \end{tabular}
\end{table}

\begin{table}[t]
  \caption{Experimental results adjusting the weight ratio of 3-class to 10-class classifications in joint training.}
  \label{tab:ratio}
  \centering
  \begin{tabular}{lcc}
    \toprule
    System & Ratio & Acc (\%) \\
    \midrule
    \multirow{5}{*}{Proposed method} & 1 : 1 & 70.3 \\
     & 1 : 2 & 69.6 \\
     & 1 : 3 & 70.3 \\
     & 1 : 4 & 70.7 \\
     & \textbf{1 : 5} & \textbf{71.3} \\
    \midrule
    GradNorm \cite{chen2018gradnorm} & - & 70.1\\
    \bottomrule
  \end{tabular}
  \vspace{-3mm}
\end{table}

\begin{table}[t]
  \caption{3-class classification accuracy. The proposed joint training also increases 3-class classification performance.}
  \label{tab:3class}
  \centering
  \begin{tabular}{lccccc}
    \toprule
    \multirow{2}{*}{System} & DCASE & Submitted sys- & Ours, w/o & Ours, w/ \\
    & Baseline& tems' average & MTL & MTL \\
    \midrule
    Acc (\%) & 88.0 & 87.3 & 91.4 & \textbf{92.2} \\
    \bottomrule
  \end{tabular}
\end{table}

\begin{table}[t]
  \caption{Comparison with recent state-of-the-art systems using the performance of individual systems without a score-level ensemble. The third to seventh rows list the top five best-performing systems on the DCASE2020 Task 1-A challenge.}
  \label{tab:sota}
  \centering
  \begin{tabular}{lccll}
    \toprule
    System & Acc (\%)  & \# Params \\
    \midrule
    Proposed Method & \textbf{71.3} & \textbf{0.6M} \\
    \midrule
    DCASE2020 Baseline \cite{heittola2020acoustic} & 54.1 & 5M \\
    Suh et al. \cite{suh2020designing} & 73.7 & 13M \\
    Hu et al. \cite{hu2020device} & 76.9 & -\\
    Gao et al. \cite{gao2020acoustic} & 71.8 & 4M \\
    Liu et al. \cite{liu2020acoustic} & 72.1 & 3M \\
    Koutini et al. \cite{koutini2020cp} & 71.8 & 225M  \\
    \bottomrule
  \end{tabular}
  \vspace{-3mm}
\end{table}

\subsubsection{Joint training}
Table \ref{tab:joint} presents the comparison of various joint training strategies.
The joint training experiments were conducted using the AMFM structure.
Both pre-training and conventional MTL did not further improve the performance. 
In the experiments, unlike~\cite{hu2020device}, joint prediction resulted in worse performance for both extended and sequential MTL frameworks. 

The best result was achieved using the extended MTL framework, which had additional layers for each training task after the last hidden layer shared by the two tasks.
The sequential MTL also demonstrated a slight performance improvement.
Through these results (original, extended, and sequential MTL), we conclude that the original MTL could not learn each task sufficiently. 
Instead, hidden layers solely assigned to solve a specific task are required. 
Based on the results, although the two tasks are similar, each output layer demands at least a few layers solely dedicated to each task. 
This interpretation is inspired by \cite{maninis2019attentive}, where the authors reported performance degradation caused by excessive interference between two related tasks. 

\subsubsection{Additional experiments}
Table \ref{tab:ratio} lists the ablation results of adjusting the weight ratio between the two subtasks for joint training.
We explored both manual and methodological approaches to determine the optimal weight ratio.
The best result was achieved when the ratio of the abstract to specific labels was 1:5.

Table~\ref{tab:3class} describes the accuracy of 3-class classification.
The MTL is effective when related tasks are jointly trained, although it is difficult to determine whether those tasks are explicitly related.
Therefore, we investigated 3-class classification performance to demonstrate that the two tasks (10-class and 3-class) are beneficial for each other.

The accuracy of the DCASE 2020 baseline of Subtask B was 88\%, and the average accuracy of the submitted system was 87.3\% \cite{heittola2020acoustic}.
When we trained the model that only performs the 3-class classification task using the same proposed AMFM structure without MTL, the accuracy was 91.4\%.
Compared with the aforementioned results, we could achieve a 92.2\% accuracy with the extended MTL scheme while demonstrating 71.3\% accuracy for the main task. 

\subsubsection{Comparison with recent state-of-the-art systems}
Table \ref{tab:sota} presents a comparison with current state-of-the-art systems without applying ensemble techniques.
The proposed system demonstrates comparable performance with state-of-the-art systems without complex data preprocessing techniques.
In addition, the proposed method was performed with low-complexity architecture.
Although it is outside the study scope, we expect that applying more data preprocessing methods might lead to further improvement.

\subsubsection{Application to the DCASE 2021 challenge}
To further demonstrate the effectiveness of the proposed AMFM, we introduce the recent result of the DCASE 2021 challenge (further details are in~\cite{Hee-Soo2021}). 
The DCASE 2021 ASC task requires lightweight systems with less than 128KB ($\approx$65k parameters in 16-bit resolution). 
Our team ranked fourth place, with two submitted systems in which one system was ResNet-based and the other was AMFM-based.
The two systems showed an accuracy of 70.1\% and 70.0\% respectively.

\section{Conclusions}

In this paper, we proposed a novel technique named AMFM that combines the merits of both the attention and MFM techniques, enabling the emphasis of discriminative information while avoiding excessive information deletion.
We also explored four joint training methods, including extended and sequential MTL frameworks.
Both proposed methods showed their effectiveness for the ASC task. 
Furthermore, the weight ratio and relevance of the two tasks in the MTL framework were experimentally investigated.  
The proposed model requires relatively few parameters compared to other state-of-the-art systems while performing competitively.
Various data preprocessing techniques adopted in the recent literature leave room for further improvement of the proposed system because both proposals in this paper focus on building a better model architecture.

\bibliographystyle{IEEEbib}
\bibliography{shortstrings,reference}
\end{document}